\documentclass[sigconf,finalstrut]{acmart}
\AtBeginDocument{%
  \providecommand\BibTeX{{%
    \normalfont B\kern-0.5em{\scshape i\kern-0.25em b}\kern-0.8em\TeX}}}

\copyrightyear{2021} 
\acmYear{2021} 
\setcopyright{acmlicensed}\acmConference[MM '21]{Proceedings of the 29th
ACM International Conference on Multimedia}{October 20--24, 2021}{Virtual
Event, China}
\acmBooktitle{Proceedings of the 29th ACM International Conference on
Multimedia (MM '21), October 20--24, 2021, Virtual Event, China}
\acmPrice{15.00}
\acmDOI{10.1145/3474085.3475290}
\acmISBN{978-1-4503-8651-7/21/10}

\usepackage{color}
\usepackage{multirow}
\usepackage{ulem}
\usepackage{fancyhdr}
\usepackage{subfig}
\begin{document}

\fancyhead{}

\title{Discriminator-Free Generative Adversarial Attack}


\author{Shaohao Lu}
\affiliation{%
  \institution{School of Computer Science and Engineering}
  \city{Guangzhou}
  \country{China}
}
\email{lushh7@mail2.sysu.edu.cn}

\author{Yuqiao Xian}
\affiliation{%
  \institution{School of Computer Science and Engineering}
  \city{Guangzhou}
  \country{China}}
\email{xianyq3@mail2.sysu.edu.cn}

\author{Ke Yan}
\affiliation{%
  \institution{Tencent Youtu Lab}
  \city{Shanghai}
  \country{China}
}
\email{kerwinyan@tencent.com}

\author{Yi Hu}
\affiliation{%
  \institution{Tencent Youtu Lab}
  \city{Shanghai}
  \country{China}
}
\email{ferdinandhu@tencent.com}

\author{Xing Sun}
\affiliation{%
  \institution{Tencent Youtu Lab}
  \city{Shanghai}
  \country{China}
}
\email{winfred.sun@gmail.com}

\author{Xiaowei Guo}
\affiliation{%
  \institution{Tencent Youtu Lab}
  \city{Shanghai}
  \country{China}
}
\email{scorpioguo@tencent.com}

\author{Feiyue Huang}
\affiliation{%
  \institution{Tencent Youtu Lab}
  \city{Shanghai}
  \country{China}
}
\email{garyhuang@tencent.com}

\author{Wei-Shi Zheng}
\affiliation{%
  \institution{School of Computer Science and Engineering}
  \city{Guangzhou}
  \country{China}}
\email{wszheng@ieee.org}

\begin{abstract}
    The Deep Neural Networks are vulnerable to \textbf{adversarial examples} (Figure \ref{fig:introduction}), making the DNNs-based systems collapsed by adding the inconspicuous perturbations to the images. Most of the existing works for adversarial attack are gradient-based and suffer from the latency efficiencies and the load on GPU memory. The generative-based adversarial attacks can get rid of this limitation, and some relative works propose the approaches based on GAN. However, suffering from the difficulty of the convergence of training a GAN, the adversarial examples have either bad attack ability or bad visual quality. In this work, we find that the discriminator could be not necessary for generative-based adversarial attack, and propose the \textbf{Symmetric Saliency-based Auto-Encoder (SSAE)} to generate the perturbations, which is composed of the saliency map module and the angle-norm disentanglement of the features module. The advantage of our proposed method lies in that it is not depending on discriminator, and uses the generative saliency map to pay more attention to label-relevant regions. The extensive experiments among the various tasks, datasets, and models demonstrate that the adversarial examples generated by SSAE not only make the widely-used models collapse, but also achieves good visual quality. The code is available at {\color{blue} \href{https://github.com/BravoLu/SSAE}{https://github.com/BravoLu/SSAE}}. 
\end{abstract}




\begin{teaserfigure}
\begin{minipage}[b]{0.48\linewidth}
\subfloat[\centering \textbf{AdvGAN} (A method \textbf{w/} discriminator.)]{
\centering
\includegraphics[width=1\textwidth, height=0.8\textwidth]{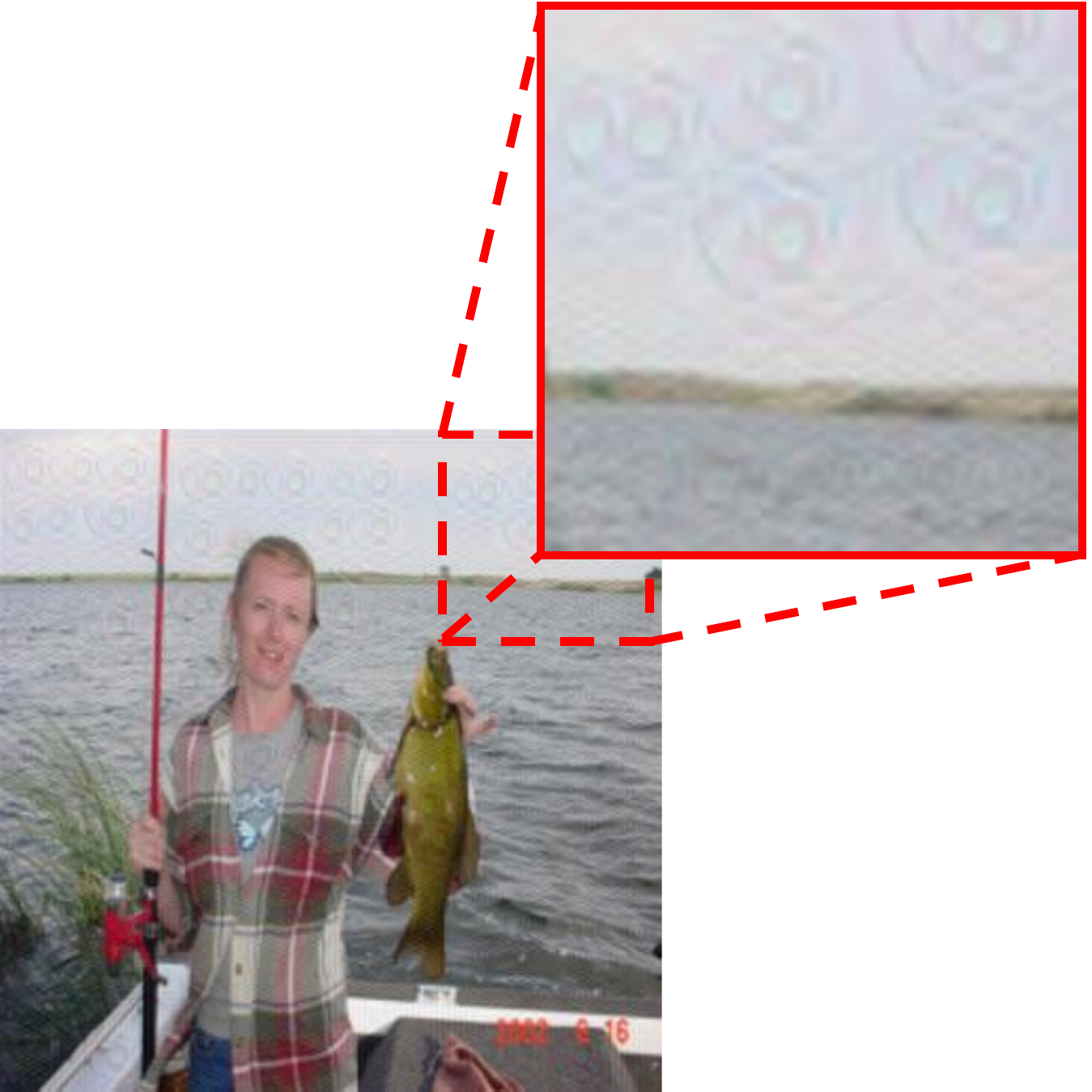}
\centering
}
\end{minipage}
\begin{minipage}[b]{0.48\linewidth}
\subfloat[\centering \textbf{Ours} (A method \textbf{w/o} discriminator.)]{
\centering
\includegraphics[width=1\textwidth, height=0.8\textwidth]{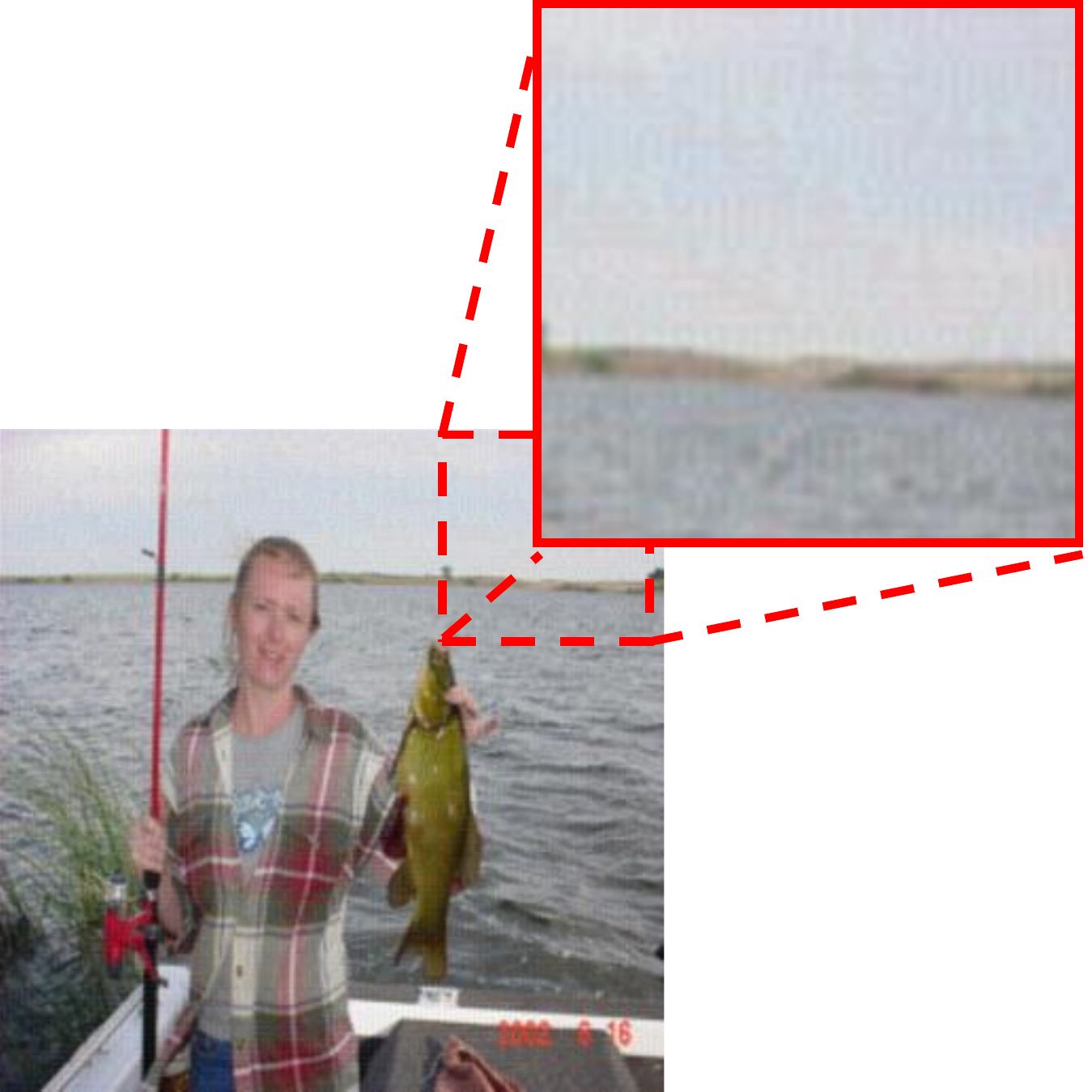}
\centering
}
\end{minipage}
\caption{Both the images above can make ResNet18 \cite{he2016deep} fail without changing in appearance significantly. The one generated by our method gets better visual quality without discriminator, while the other one generated by AdvGAN \cite{ijcai2018-543} has conspicuous circular noises.}
\label{fig:introduction}
\end{teaserfigure}

\maketitle

\section{Introduction}

The Deep Neural Networks(DNNs) play an increasingly important role on a wide variety of computer vision tasks, e.g, object classification, object detection, and image retrieval. Despite the breakthroughs brought by the DNNs for these tasks, the DNNs are found to be vulnerable to the adversarial examples \cite{szegedy2013intriguing,papernot2016transferability,carlini2017towards,papernot2016limitations,chen2018ead}, 
which brings the severe security risk in the safety-critical applications. Szegedy et.al  \cite{szegedy2013intriguing} firstly proposed that the adversarial examples can make the classifier based on the DNNs collapsed, which is indistinguishable to the human visual systems. 
The lack of robustness of the DNNs to the adversarial examples has raised serious concerns for the security of the DNN-based systems closely related to human safety, e.g., the autonomous driving system. 

Prevalent adversarial attack methods \cite{carlini2017towards,papernot2016limitations,chen2018ead} are gradient-based, which have been proven for their effectiveness in attacking the models for classification. Su et.al \cite{su2019one} successfully attacks the images by modifying just one pixel. However, the limitation of these methods comes from its latency efficiencies, specifically, it needs multiple times of the gradient back-propagation to obtain the adversarial examples. To tackle this limitation, Xiao et.al \cite{xiao2018generating} proposed a method based on GAN called AdvGAN, which can instantly produce adversarial examples. However, the GAN-based adversarial network always suffers from the convergence because of the min-max game between the generator and the discriminator \cite{arjovsky2017wasserstein}, and always lead to either the perceptible perturbation or bad attack ability. Training a effective GAN needs strict training strategy. To address the aforementioned issues, we propose a discriminator-free network: Symmetric Saliency-Based Auto-Encoder, to generate the more effective and inconspicuous perturbations. 

Different from the existing attack methods which target at confusing the decision boundary, our method attacks the target model on the feature space, which guarantees its transferability among various tasks. Specifically, we propose angle-norm disentanglement to disentangle the cosine angle and norm between the features of raw images and those of the perturbed images, and push away their angles to make the systems misidentify the perturbed images. In addition, to improve the inconspicuousness of the adversarial examples, we propose a saliency map module to restrict the size of the target region, and weaken the perturbations in label-irrelevant regions of images (e.g., the background, the label-irrelevant objects), while making the model pay more attention to the label-relevant areas (i.e., the main object). 

To validate the effectiveness of the proposed method, we conduct extensive experiments on two tasks: (1) Image classification and (2) image retrieval (person re-identification). The experimental results validate the superiority of our method, ranging from the attack ability, the visual quality, and the transferability in the semi-black box (transfer) manner.

In summary, this work proposes a novel adversarial attack method for object recognition. The contributions include the following:
\begin{itemize}
    \item We consider that attacking the corresponding systems without discriminator is more simplified but comparably effective for adversarial attack. We consider that only the magnitude of the perturbation is tiny enough, the adversarial examples can deceive the human eyes. It is not necessary that whether the discriminator can distinguish the adversarial examples or not. We conduct lots of analysis and experiments for this perspective.
    \item We propose a novel discriminator-free generative adversarial attack network called SSAE. The saliency map module of it makes the perturbations more deceptive while keeping the attack ability.
    \item The angle-norm disentanglement of SSAE is proposed to attack the target model by pushing away the angles between the raw instances and the perturbed instances, which enables our method to be easily applied in various tasks with limited modification. We successfully attack the image classification systems and image retrieval systems, and conduct lots of experiments about transferablity.
\end{itemize}

\section{Related Work}
\noindent {\bf Adversarial Attack.}
According to the goal of the adversarial attack, it can be classified into two categories: the \textit{targeted adversarial attack} and the \textit{non-targeted adversarial attack}. The targeted attack requires the target model to recognize the adversarial attack as the specific category, while the non-targeted attack only focuses on whether the adversarial examples deceive the target model or not. According to the degree of acquisition to the target model, adversarial attack also can be classified into another two categories: the \textit{white-box adversarial attack} and the \textit{black-box adversarial attack}. In the white-box adversarial attack, we have complete access to the target model, including the architecture of the neural network and the weights of the neural network. With this information about the target model, we can use a gradient-based method to obtain adversarial examples. The most existing adversarial attack methods are in a white-box manner. The black-box attack is a more challenging task, where we only have the output of the target model. Our method is mainly in the non-target and the white-box manner, and we also test its performance in the black-box (transfer) manner.

~\\
\noindent {\bf Adversarial Attack for Image Classification.} Most existing adversarial attack methods \cite{madry2018towards,carlini2017towards,papernot2016limitations,chen2018ead} are proposed for the image classification task. Szegedy \cite{szegedy2013intriguing} firstly proposed the adversarial examples by formulating the following minimization problem.
\begin{equation}
    \arg\min_{\mathcal{P}} f(I+\mathcal{P}) \ne f(I) \quad s.t.\ (I+\mathcal{P})\in D.
\label{eq.formula}
\end{equation}
The input example $I$ will be misclassified after being perturbed by the perturbation $\mathcal{P}$, and $I+\mathcal{P}$ are constrained in the domain $D$. For keeping the inconspicuousness of adversarial example, the $\mathcal{P}$ is expected to be minimized. The \textit{L-BFGS} \cite{liu1989limited} are used to solve this equation for non-convex models like DNN. Goodfellow et.al. \cite{goodfellow2014explaining} proposed the \textit{FGSM}, a more efficient solution to solve Eq (\ref{eq.formula}). In addition, the \textit{PGD} \cite{madry2018towards}, the \textit{C\&W} \cite{carlini2017towards} and the \textit{JSMA} \cite{papernot2016limitations} also have achieved great success on the adversarial attacks on image classification. All the aforementioned methods are effective to obtain good adversarial examples but suffer from their expensive computation. For each input image, these approaches search for the optimal perturbation $\mathcal{P}$ with the gradients, which is a time-consuming process. The GAN-based method \cite{xiao2018generating} is proposed to alleviate this limitation, which only needs forward inference once instead of iterative computations. However, the limitation of GAN-based methods lies in training a good generative adversarial network.

~\\
\noindent {\bf Adversarial Attack for Image Retrieval.} Adversarial Attack for Image Retrieval is a more challenging task because, in the image retrieval task, a query image searches through a database on visual features in the absence of the class labels. Recently, some works \cite{liu2019s,li2019universal,zheng2018open,tolias2019targeted} target in the adversarial attack for image retrieval, all of them are gradient-based approaches. Wang et.al \cite{Wang_2020_CVPR} firstly proposed a GAN-based adversarial attack approach for person re-identification, a specific task of image retrieval. It has a high attack success rate for ReID systems, but poor visual quality. We assume that suffering from the min-max game between generator and discriminator, the perturbations generated by GAN are not stable, thus leading to poor visual quality.

\section{Symmetric Saliency-Based Auto-Encoder}
\begin{figure*}[!t]
\begin{center}
\includegraphics[width=1\linewidth]{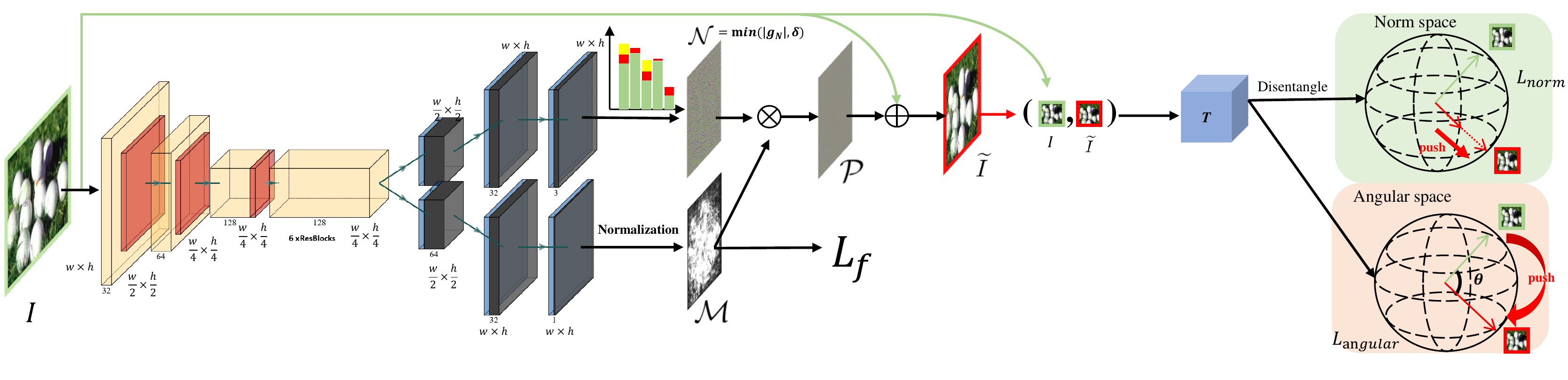}
\end{center}
   \caption{The overview of the Symmetric Saliency-based Auto-Encoder for adversarial attack. The original images $I$ are feed forward the auto-encoder to generate the perturbations $\mathcal{N}_0$ and saliency map $\mathcal{M}$. The raw image and the perturbed image pairs $(I,\widetilde{I})$ are feed into the target model $\mathcal{T}$, and disentangled with the angles and the norms. The angle-norm disentanglement pushes away the angles between $I$ and $\widetilde{I}$ to collapse the target model while keeps the norms changeless for better visual quality. }
\label{fig:framework}
\end{figure*}
\subsection{Overview}
Figure \ref{fig:framework} illustrates the proposed Symmetric Saliency-Based Auto-Encoder framework for adversarial attack. Compared with the GAN-based adversarial attack methods, which are always limited by its convergence difficulty because of the min-max games between the generator and the discriminator, our method based on auto-encoder can obtain adversarial examples with both more good attack ability and visual quality. The symmetric saliency-based auto-encoder consists of an encoder $\mathcal{E}$ and two decoders. Specifically, the two decoders are symmetric: (1) The noise decoder $\mathcal{G}_\mathcal{N}$ and (2) the saliency map decoder $\mathcal{G}_\mathcal{M}$. The original image $I$ is encoded as the latent vectors by the encoder. The latent vectors are then decoded by noise decoder $\mathcal{G}_\mathcal{N}$ to generate noise $\mathcal{N}$, and decoded by saliency map decoder $\mathcal{G}_\mathcal{M}$ to generate the mask $\mathcal{M}$, respectively. The $\mathcal{M}$ is designed for alleviating the magnitude of the perturbation in the label-irrelevant region for better visual quality. We denote the raw instance as $I$, the perturbed instance as $\widetilde{I}$, and the features before the fully connected layers of the target model as $\mathcal{T}(\cdot)$. For making the target models misidentify the adversarial examples, we conduct the angle-norm disentanglement which pushes away the cosine angles between $\mathcal{T}(I)$ and $\mathcal{T}(\widetilde{I})$ for mismatching while keeping the value of their norms changeless.

\subsection{The Discriminator-Free Adversarial Attack}
Most existing generative-based methods of adversarial attack are of a similar GAN-based architecture, which consists of a generator and a discriminator. The discriminator $\mathcal{D}$ are supposed to distinguish the adversarial ones or the original ones, to keep the visual inconspicuousness of perturbed instances. However, we consider that the discriminator may be redundant in the adversarial attack as long as the perturbations are tiny enough. The min-max games between generator and discriminator only guarantee the indistinguishability between original ones and perturbed ones in the discriminative space rather than human visual systems.

The min-max games make the algorithm hard to converge, which needs strict training strategy, thus usually leading to either bad visual quality or bad attack ability. Based on the above assumptions, we propose a novel network to get rid of the difficulties of training a GAN, which is simplified but more effective. It is a discriminator-free and generative-based adversarial attack method. The overview of symmetric saliency-based auto-encoder is illustrated in Figure \ref{fig:framework}. The network consists of three modules: (1) The encoder (2) the perturbation decoder and (3) the saliency map decoder. 

\noindent \textbf{The encoder $\mathcal{E}$} is lightweight, which is composed of one $7\times7$ convolution, two $3\times3$ convolutions and six \textit{ResBlocks} \cite{he2016deep}.

\noindent \textbf{The perturbation decoder $\mathcal{G}_\mathcal{N}$} consists of two $3\times3$ transposed convolutions and one $7\times7$ transposed convolution. The dimension of the output is $3\times H\times W$, the same as that of the input. However, the raw perturbations $\mathcal{N}_0$ may be too large to deceive the human visual system, so we make an extra constraint for its value:
\begin{equation}
    \mathcal{N}(I) = min(|{\mathcal{G}}_{\mathcal{N}}(I)|, \delta),
\label{eq:delta}
\end{equation}
we can manually set the $\delta$ according to our needs: the larger $\delta$ means the higher attacking performance while the smaller one means the better visual quality. In our experiments, we set $\delta=0.1$.

\subsection{The Saliency Map for Effective Attack}
\noindent \textbf{The saliency map decoder $\mathcal{G}_\mathcal{M}$} has a similar architecture with the perturbation decoder except the last transposed convolution layer. It is obvious that some regions of the images are critical for the DNN to recognize (i.e., the label-relevant objects), while some regions of the images are less important (e.g., background). Adding the perturbation to label-relevant regions of images leads to a more effective attack. A simple idea is that we can use Mask-RCNN \cite{he2017mask} to get the corresponding foreground masks. However, there are some defects of directly applying object segmentation algorithms for generating masks: (1) The obvious boundary of the attack region sometimes leads to the worse visual quality, which makes the perturbation more obvious. (2) The foreground mask can not reflect the relative importance of a specific region of the foreground. To tackle this problem, we propose a saliency map decoder to learn the relative importance of the region by itself. The dimension of the output of $\mathcal{G}_\mathcal{M}$ is $1\times W\times H$. We obtain the normalized matrix $\mathcal{M}$ by:
\begin{equation}
    \mathcal{M}(I^{r,c}) = \frac{\mathcal{G}_\mathcal{M}(I^{r,c}) - \min\limits_{r,c}(\mathcal{G}_\mathcal{M}(I^{r,c}))}{\max\limits_{r,c}(\mathcal{G}_\mathcal{M}(I^{r,c}) - \min\limits_{r,c}(\mathcal{G}_\mathcal{M}(I^{r,c})))},
\end{equation}
such that the value of $\mathcal{M}(I^{r,c})$ can represent the importance of the point in location $(r,c)$ of the image $I$. In order to make the attacked regions more concentrated, we use the Frobenius norm to optimize the parameters of the saliency map decoder, which makes the saliency map more concentrated. The formula for it is as follows:
\begin{equation}
    \mathcal{L}_f = \sum_{i=1}^{N}\sqrt{tr(\mathcal{M}(I_i)^T\mathcal{M}(I_i))},
\label{eq:fn}
\end{equation}
where the $tr(\cdot)$ is the trace function, and the $N$ is the number of images. The Eq.(\ref{eq:fn}) is out of any label supervision, so the label-relevant regions can be well extracted only with other objectives like Eq.(\ref{eq:cos}).

\subsection{The Angle-Norm Disentanglement for Mis-matching}

\begin{figure}[!t]
\begin{minipage}[b]{0.45\linewidth}
\subfloat[\textbf{The original distribution of the CNN features.}]{\includegraphics[width=1\textwidth, height=0.8\textwidth]{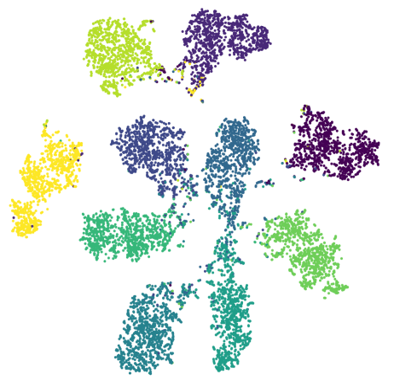}}
\end{minipage}
\begin{minipage}[b]{0.45\linewidth}
\subfloat[\textbf{The perturbed distribution of the CNN features.}]{\includegraphics[width=1\textwidth, height=0.8\textwidth]{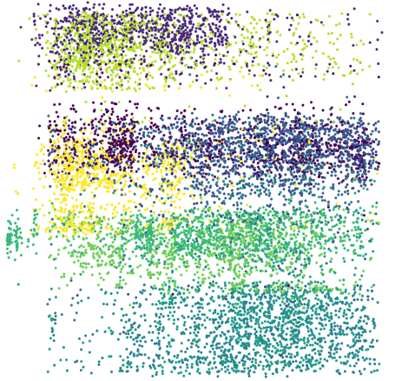}}
\end{minipage}
\caption{The t-SNE visualization of the CNN features of \textit{cifar-10}. The distribution totally changes after pushing way their angles.}
\label{fig:tsne}
\end{figure}

Chen et.al \cite{chen2019angular} revealed the gap between the human visual system and the DNN models, and found that the human selection frequency is highly correlated to the angle of the features. Chen et.al \cite{chen2020norm} proposes norm-aware embedding for efficient person search. We visualize the distribution of the CNN features of all the images in the test set of \textit{cifar-10} with t-SNE \cite{van2008visualizing} in Figure \ref{fig:tsne} for validation, and we can find that the distribution of the CNN features of \textit{cifar-10} highly depends on the angle. Inspired by the aforementioned facts, we disentangle the norm and angles from the original instance and perturbed instance respectively, as shown in the right part of Figure \ref{fig:framework}. 

\noindent \textbf{Angle}. For effectively attacking the target models, we focus on applying an attack on the angle between perturbed instance and the original instance. We attempt to push away the cosine angle between the CNN feature of the original instance and the perturbed instance using the following objective formulas:
\begin{equation}
    \mathcal{P}(I_i)  =  \mathcal{N}(I_i)  \odot \mathcal{M}(I_i), 
\end{equation}
\begin{equation}
       \mathcal{L}_{angular} = \sum_{i=1}^{N}(1+\frac{\mathcal{T}(I_i)\cdot \mathcal{T}(I_i+\mathcal{P}(I_i))}{max(||\mathcal{T}(I_i)|| \cdot ||\mathcal{T}(I_i+\mathcal{P}(I_i))||, \epsilon)}), 
\label{eq:cos}
\end{equation}
where the $\mathcal{P}$ is the perturbation and the $\epsilon$ is set as $1e-12$ for numeric stability. 

\noindent \textbf{Norm}. We assume that the less we change the features, the more deceptive the perturbed instances are. So we keep the value of the norm changeless as much as possible for better visual consistency using the following objective formulas:
\begin{equation}
    \mathcal{L}_{norm} = \sum^{N}_{i=1}(||\mathcal{T}(I_i)||-||\mathcal{T}(I_i+\mathcal{P}(I_i))||)^2.
\end{equation}
The overall loss objective is:
\begin{equation}
    \mathcal{L} = \mathcal{L}_{angular} + \alpha (\mathcal{L}_{norm} + \mathcal{L}_{f}),
\label{eq:objective}
\end{equation}
where $\alpha$ is the hyperparameter that controls the relative importance of the $\mathcal{L}_{norm}$ and the $\mathcal{L}_{f}$. Finally, we can obtain the perturbed instance by:
\begin{equation}
    \widetilde{I} = tensor2img(I + \mathcal{P}(I)),
\end{equation}
where the $tensor2img(\cdot)$ is the inverse transformation of normalization.

\section{Experiments}

\begin{table*}[!htbp]
\small
\caption{The performance of attacking the classification networks. Our method achieves comparable results on both the attack ability and the visual quality. The AdvGAN does not work in ResNet18, EfficientNet-B0, and MobileNetv2 at all for \textit{imagenette} ({\color{red}red}). We apply 20*log to psnr. ResNet18:\cite{he2016deep}, EfficientNet-B0:\cite{tan2019efficientnet}, GoogLeNet:\cite{szegedy2015going}, MobileNetv2\cite{sandler2018mobilenetv2}, DenseNet121\cite{huang2017densely}.} 
\resizebox{\textwidth}{42mm}{
\begin{tabular}{l|lllll|llll}
\multicolumn{10}{c}{(a) \textbf{\textit{cifar10}}}                                                                                                            \\ 
\hline
\hline
\multirow{2}{*}{\textbf{Models}} & \multicolumn{5}{c|}{\textbf{Accuracy(\%)}} & \multicolumn{4}{c}{\textbf{ssim/ms-ssim/psnr}} \\ \cline{2-10} 
                       & before & PGD & FGSM  & AdvGAN  & Ours    & PGD & FGSM &  AdvGAN       & Ours \\ \hline
ResNet18               & 95.4  &  28.0 & 49.1 & 20.2  & \textbf{8.2}  & 1.0/1.0/97.0 & 0.96/1.0/82.1 &  0.95/1.0/69.7        & 0.96/1.0/82.2 \\
EfficientNet-B0           & 91.2  & 23.8 & 10.2  & 20.5   & \textbf{11.4} & 1.0/1.0/96.9   &  0.96/1.0/82.1      &0.94/0.99/69.7      & 0.96/1.0/82.4\\ 
GoogLeNet              & 95.2 & \textbf{12.7} & 26.3  & 43.8  & 19.8 & 1.0/1.0/97.0       & 0.96/1.0/82.1      & 0.95/1.0/69.3         & 0.97/1.0/82.5\\ 
MobileNetv2            & 97.0 & 26.6 & 18.6   &52.7  & \textbf{4.1}  & 1.0/1.0/97.0        & 0.96/1.0/82.1      & 0.95/1.0/69.3        & 0.96/1.0/82.4\\ 
DenseNet121               & 95.6  & 21.1 & 42.4 & 19.1  & \textbf{7.1}  & 1.0/1.0/97.0      & 0.96/1.0/82.1    & 0.95/1.0/69.2          & 0.97/1.0/82.4 \\ 
\hline
\hline
\multicolumn{10}{c}{(b) \textbf{\textit{imagenette}}}                                                                                                         \\ 
\hline
\hline
\multirow{2}{*}{\textbf{Model}} & \multicolumn{5}{c|}{\textbf{Accuracy(\%)}} & \multicolumn{4}{c}{\textbf{ssim/ms-ssim/psnr}} \\ \cline{2-10} 
                       & before & PGD & FGSM  & AdvGAN  & Ours    & PGD & FGSM &  AdvGAN       & Ours \\ \hline
ResNet18               & 94.0  &   78.4 & \textbf{8.9} & {\color{red} 94.0}  & 11.1  & 1.0/1.0/114.1  & 0.87/0.98/82.1 &  {\color{red} 1.0/1.0/149.7}        & 0.95/1.0/86.2 \\
EfficientNet-B0           & 91.1  & 77.2 & \textbf{7.2}  & {\color{red} 91.1}   & 8.5 & 1.0/1.0/114.1   & 0.87/0.87/82.1      & {\color{red} 1.0/1.0/157.3}     & 0.94/0.99/85.4\\ 
GoogLeNet              & 90.4 & 63.6 & 14.1  & 16.0  & \textbf{12.1} & 1.0/1.0/114.0       & 0.87/0.98/82.1      & 0.91/0.99/84.0         & 0.94/1.0/86.4\\ 
MobileNetv2            & 93.7 & 49.0 & \textbf{8.5}   &{\color{red} 93.7}  & 10.7  & 1.0/1.0/114.0        & 0.87/0.98/82.1      & {\color{red} 1.0/1.0/156.0}        & 0.95/1.0/86.7\\ 
DenseNet121               & 86.3  & 70.4 & 22.5 & \textbf{10.8}  & 19.2  & 1.0/1.0/114.0      & 0.88/0.97/82.1    & 0.90/0.99/83.8          & 0.97/1.0/89.5 \\ 
\hline
\hline
\multicolumn{10}{c}{(c) \textbf{\textit{caltech101}}}                                                                                                         \\ 

\hline
\hline
\multirow{2}{*}{\textbf{Models}} & \multicolumn{5}{c|}{\textbf{Accuracy(\%)}} & \multicolumn{4}{c}{\textbf{ssim/ms-ssim/psnr}} \\ \cline{2-10} 
                       & before & PGD & FGSM  & AdvGAN  & Ours    & PGD & FGSM &  AdvGAN       & Ours \\ \hline
ResNet18               & 72.1  &  64.5 & \textbf{14.1} & 14.3  & 14.8  & 0.99/0.99/113.9 & 0.86/0.97/82.1  &  0.64/0.96/74.6        & 0.86/0.98/82.3 \\
EfficientNet-B0           & 76.1  & 71.5   & \textbf{10.2}   & 31.2 & 14.4 & 0.99/0.99/113.9  &  0.86/0.97/82.1       & 0.64/0.96/74.5      & 0.87/0.97/82.3\\ 
GoogLeNet              & 57.9 & 36.2 & 11.6  & 14.9  & \textbf{6.7} & 0.99/0.99/113.9       & 0.86/0.98/82.1      &0.61/0.95/74.3         & 0.88/0.98/82.7\\ 
MobileNetv2            & 69.4 & 47.3 & 10.0   &27.4  & \textbf{3.1}  & 0.99/0.99/113.9        & 0.86/0.98/82.1      & 0.63/0.95/74.7        & 0.87/0.97/82.5\\ 
DenseNet121               & 54.9  & 47.3 & 15.6 & 9.5  & \textbf{2.8}  & 0.99/0.99/113.9      & 0.86/0.98/82.1    & 0.68/0.96/74.0          & 0.86/0.99/82.3 \\ 
\hline
\hline

\end{tabular}
}
\label{tab:classifiation_result}
\end{table*}

\begin{table*}[!t]
\caption{The performance of attacking the various ReID systems. Compared to the GAN-based method DMR\cite{Wang_2020_CVPR}, our method outperforms in all the three classical metrics that applied in the fully reference of image quality assessment. We apply 20*log to psnr. IDE:\cite{zheng2016person}, Mudeep:\cite{qian2017multi}, AlignedReID:\cite{zhang2017alignedreid}, PCB:\cite{sun2018beyond}, HACNN:\cite{li2018harmonious}, CamStyle-Era:\cite{zhong2018camera}, HHL:\cite{zhong2018generalizing}, SPGAN:\cite{deng2018image}.}
\centering
\resizebox{\textwidth}{30mm}{
\begin{tabular}{c|l|lllll|lllll|ll}
\multicolumn{14}{c}{(a) \textbf{\textit{Market1501}}}                                                                                                                                                    \\
\hline
\hline
\multicolumn{2}{c|}{\multirow{2}{*}{\textbf{Methods}}}              & \multicolumn{5}{c|}{\textbf{Rank-1 (\%)}}             & \multicolumn{5}{c|}{\textbf{mAP (\%)}}               & \multicolumn{2}{c}{\textbf{ssim/ms-ssim/psnr}} \\
\cline{3-14}
\multicolumn{2}{l|}{}                                      & Before & GAP  & PGD  & DMR   & Ours & Before & GAP & PGD  &DMR   & Ours & DMR           & Ours              \\
\cline{1-14}
\multirow{2}{*}{Backbone}          & IDE(ResNet50)        & 83.1   & 5.0  & 4.5  & 3.7 & \textbf{2.1}  & 63.3   & 5.0 & 4.6  & 4.4 & \textbf{2.6}  & 0.69/0.89/73.0 & \textbf{0.92/0.98/83.1} \\
                                   & Mudeep(Inception-V3) & 73.0   & 3.5  & 2.6  & \textbf{1.7} & 8.4  & 49.9   & 2.8 & 2.0  & \textbf{1.8} & 5.7  & 0.62/0.90/73.2 & \textbf{0.97/0.99/82.3} \\
\cline{1-14}
\multirow{3}{*}{Part-Aligned}      & AlignedReID          & 91.8   & 10.1 & 10.2 & 1.4 & \textbf{0.9}  & 79.1   & 9.7 & 8.9  & 2.3 & \textbf{2.2}  & 0.56/0.90/73.0 & \textbf{0.93/0.98/83.7} \\
                                   & PCB                  & 88.6   & 6.8  & 6.1  & \textbf{5.0} & 9.2  & 70.7   & 5.6 & 4.8  & \textbf{4.3} & 6.0  & 0.78/0.91/72.9 & \textbf{0.94/0.97/81.6} \\
                                   & HACNN                & 90.6   & 2.3  & 6.1  & \textbf{0.9} & 5.5  & 75.3   & 3.0 & 5.3  & \textbf{1.5} & 5.1  & 0.72/0.95/73.0 & \textbf{0.95/0.99/81.5} \\
\cline{1-14}
\multirow{3}{*}{Data Augmentation} & CamStyle-Era (IDE)   & 86.6   & 6.9  & 15.4 & 3.9 & \textbf{2.7}  & 70.8   & 6.3 & 12.6 & 4.2 & \textbf{3.3}  & 0.61/0.90/73.1 & \textbf{0.93/0.98/82.8} \\
                                   & HHL(IDE)             & 82.3   & 5.0  & 5.7  & 3.6 & \textbf{1.5}  & 64.3   & 5.4 & 5.5  & 4.1 & \textbf{2.5}  & 0.71/0.91/72.9 & \textbf{0.93/0.98/83.3} \\
                                   & SPGAN(IDE)           & 84.3   & 8.8  & 10.1 & \textbf{1.5} & 6.2  & 66.6   & 8.0 & 8.6  & \textbf{1.6} & 5.1  & 0.71/0.93/72.9 & \textbf{0.94/0.98/82.6} \\
\hline
\hline


\multicolumn{14}{c}{(b) \textbf{\textit{cuhk03}}}                                                                                                                                        \\
\hline
\hline
\multicolumn{2}{c|}{\multirow{2}{*}{\textbf{Methods}}}   & \multicolumn{5}{c}{\textbf{Rank-1 (\%)}}           & \multicolumn{5}{c}{\textbf{mAP (\%)}}             & \multicolumn{2}{c}{\textbf{ssim/ms-ssim/psnr}} \\
\cline{3-14}
\multicolumn{2}{l|}{}                          & Before & GAP & PGD & DMR   & Ours & Before & GAP & PGD &DMR   & Our &DMR           & Ours              \\
\cline{1-14}
\multirow{2}{*}{Backbone}     & IDE(ResNet50) & 24.9   & 0.9 & 0.8 & \textbf{0.4} & 0.6  & 24.5  & 2.0 & 6.7 & \textbf{0.7} & 0.9 & 0.78/0.90/73.4 & \textbf{0.96/0.98/84.9} \\
                              & Mudeep        & 32.1   & 1.1 & 0.4 & 0.1 & 0.6  &          30.1   & 3.7 & 1.0 & \textbf{0.5} & \textbf{0.5} & 0.75/0.96/73.8 & \textbf{0.96/0.98/83.2} \\
\cline{1-14}
\multirow{3}{*}{Part-Aligned} & AlignedReID   & 61.5   & 2.1 & 1.4 & \textbf{1.4} & 2.4  & 59.6   & 4.6 & 2.2 & 3.7 & \textbf{2.7} & 0.68/0.92/73.2 & \textbf{0.97/0.98/85.0} \\
                              & PCB           & 50.6   & 0.9 & 0.5 & \textbf{0.2} & 2.6  & 48.6   & 4.5 & 2.1 & \textbf{1.3} & 2.4 & 0.73/0.86/73.0 & \textbf{0.93/0.96/81.5} \\
                              & HACNN         & 48.0   & 0.9 & 0.4 & \textbf{0.1} & 2.6  & 47.6   & 2.4 & 0.9 & \textbf{0.3} & 2.8 & 0.92/0.93/73.1 & \textbf{0.96/0.98/81.9} \\
\hline
\hline
\end{tabular}
}
\label{tab:ir}
\end{table*}

\begin{table*}[!t]
\small
\centering
\caption{The time consumption and the GPU memory consumption of the generative-based attack methods and the gradient-based methods. The experimental results are conducted on the imagenette, GPU GTX1080, pytorch and $batch size=1$.}
\resizebox{\textwidth}{12mm}{
\begin{tabular}{c|l|lllll|lllll}
\hline
\hline
\multirow{2}{*}{\textbf{Category}} & \multirow{2}{*}{\textbf{Methods}} & \multicolumn{5}{c|}{\textbf{Speed (FPS)}}                              & \multicolumn{5}{c}{\textbf{GPU Memory (MiB)}}                         \\ 
\cline{3-12} 
\multicolumn{2}{l|}{} & ResNet18 & EfficientNet-B0 & GoogleNet & MobileNetV2 & DenseNet121 & ResNet18 & EfficientNet-B0 & GoogleNet & MobileNetV2 & DenseNet121 \\ 
\hline
\multirow{2}{*}{generative}   & AdvGAN  & 60.9        & 42.4            & 16.1         & 36.3           & 16.0        & 741        & 675            & 1043         & 753           & 1251        \\ 
                              & Ours    & 27.1        & 21.4            & 12.4         & 20.0           & 11.7        & 773        & 703            & 1039         & 795           & 1283        \\ 
\hline
\multirow{2}{*}{gradient}    & PGD     & 0.43        & 0.34            & 0.10         & 0.25           & 0.08        & 1437        & 1561            & 5167        & 2559          & 6461        \\ 
                              & FGSM    & 11.53        & 7.7            & 2.46         & 6.60           & 1.56       & 1579       & 1435            & 5165         & 2587           & 6559        \\ 
\hline
\hline
\end{tabular}}
\label{tab:consumption}
\end{table*}

\subsection{Attack On Image Classification}
The most existing adversarial attacks on the image classification are performed on \textit{cifar-10}. However, the images of \textit{cifar-10} are too small to be aware of the perturbations. For better evaluating the inconspicuousness of perturbation, we additionally perform our attack on two more datasets. One is a dataset provided by \textit{fast.ai}: \textit{imagenette} \cite{imagewang}, which is a subset of ImageNet, composed of 10 classes, and the other is dataset \textit{caltech101} \cite{fei2004learning}. We randomly choose 20 images as the test set for \textit{caltech101}. To fully evaluate the effect of the proposed attack, we select five representative classification networks to attack, including ResNet18 \cite{he2016deep}, EfficientNet-B0 \cite{tan2019efficientnet}, GoogLeNet \cite{szegedy2015going}, MobileNetV2 \cite{sandler2018mobilenetv2} and DenseNet121 \cite{huang2017densely}. For quantitatively evaluating the overall inconspicuousness of our attack, we refer to the three classical metrics that widely used in thse full reference of Image Quality Assessment (IQA): the structural similarity index measure (\textbf{ssim}) \cite{wang2004image}, the multi-scale structural similarity (\textbf{ms-ssim}) \cite{wang2003multiscale} and the peak-signal-to-noise ratio (\textbf{psnr}) for references. The experimental results are illustrated on Table \ref{tab:classifiation_result}. 

\noindent \textbf{Attack ability.} The results show that our attack can 
make all the tested classifiers unreliable, for example, our attack makes the ResNet18 obtain 87.2\%/82.9\% accuracy degradation on the \textit{cifar-10} and the \textit{imagenette}, respectively. 
Under the same setting ($\delta=0.1)$, the PGD obtains the best visual quality while the worst attack ability. Specifically, the AdvGAN does not work in lightweight networks (ResNet18, EfficientNet-B0, and MobileNetv2) in \textit{imagenette} at all. To some extent, it validates that our assumption: the GAN-based adversarial attacks are not stable.

\begin{figure}[!t]
\subfloat[\textbf{The original images.}]{
\begin{minipage}[b]{1\linewidth}
\includegraphics[width=1\textwidth, height=0.23\textwidth]{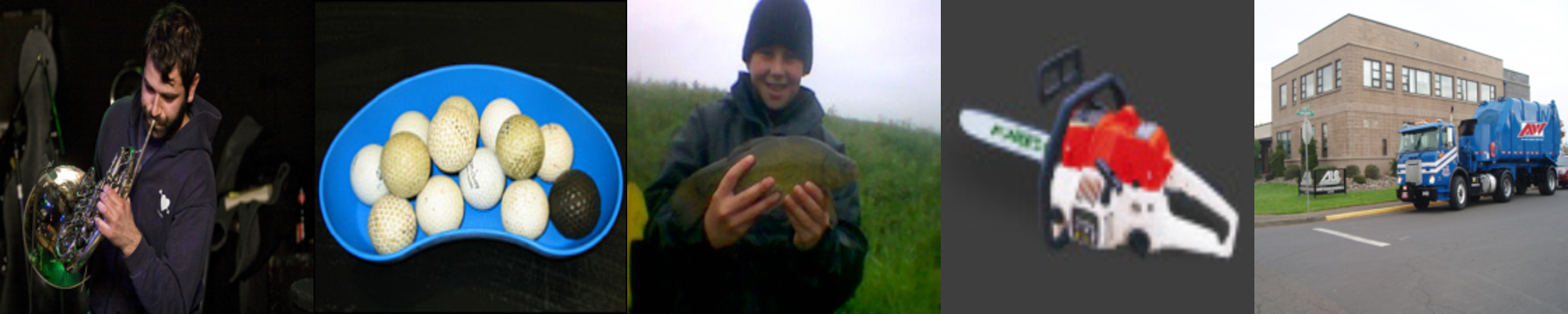}
\end{minipage}
}

\subfloat[\textbf{The heat maps {\color{green}before} attack.}]{
\begin{minipage}[b]{1\linewidth}
\includegraphics[width=1\textwidth, height=0.23\textwidth]{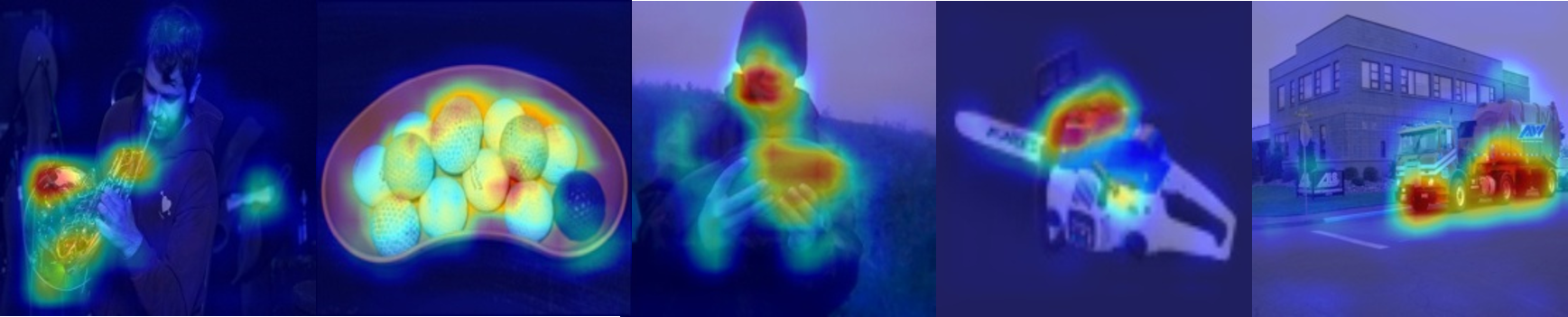}
\end{minipage}
}

\subfloat[(c) \textbf{The heat maps {\color{red}after} attack.}]{
\begin{minipage}[b]{1\linewidth}
\includegraphics[width=1\textwidth, height=0.23\textwidth]{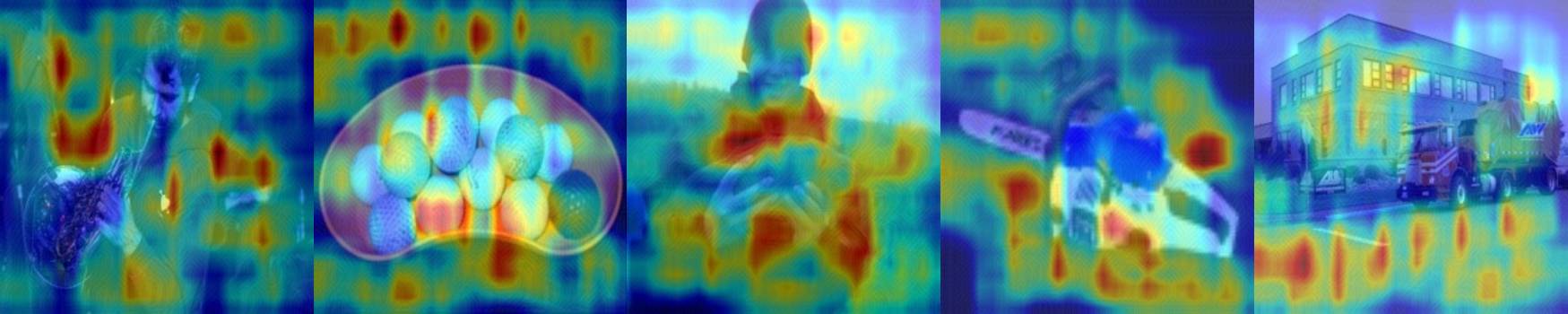}
\end{minipage}
}
\caption{The heat maps generated with Grad-cam\cite{selvaraju2017grad} are totally changed after attacked by our proposed method. (\textit{imagenette})}
\label{fig:cam}
\end{figure}

\noindent \textbf{Visual quality.} A typical adversarial sample of the \textit{imagenette} generated by our method are illustrated in Figure \ref{fig:introduction}. More adversarial examples of \textit{caltech101} generated by our method are provided in supplementary for reference. We find that the predicted categories of the models are changed though the visual performance changes little. We also use the Grad-cam \cite{selvaraju2017grad} to visualize the label-relevant regions before and after the attack in Figure \ref{fig:cam}. The results show that the label-relevant regions change totally after being attacked by our proposed method.

\subsection{Attack On Image Retrieval}
We also validate
the effectiveness of our model on Person Re-ID, a fine-grained image retrieval task aiming at matching pedestrians across non-overlapping cameras. Wang et.al proposed a GAN-based adversarial attacks (denoted as \textbf{DMR} \cite{Wang_2020_CVPR} in the following) on person re-identification, but we found that the perturbations of DMR are easy to detect. One possible reason might be the instability of the GAN-based method. To have a fair comparison with the DMR, we conduct the corresponding experiments with the same protocols as the DMR: the same target models with the same weights, and the same datasets. According to the protocols in the DMR, we attack the model from three aspects: the backbones, the part-aligned model and the data augmentation on the two widely-used datasets: The \textit{Market1501} \cite{zheng2015scalable} and the \textit{cuhk03} \cite{li2014deepreid}, to validate the robustness of our method. The results are illustrated in Table \ref{tab:ir}. 

\noindent \textbf{Attacking ReID Systems with Different Backbones.}
We firstly attack the models with the different backbones: The ResNet50 and the Inception-V3. Both the rank-1 and mAP accuracy drop sharply.

\noindent \textbf{Attacking the Part-based ReID Systems.} 
Many efficient ReID systems are based on partial alignment. However, the results tell that our attacks also can collapse these systems. For instance, the AlignedReID model gets about 77\% degradation in mAP after being attacked, and the PCB can not resist our attack, either.   

\noindent \textbf{Attacking Augmented ReID Systems.} 
Data augmentation is a widely-used technique to increase the generalization performance of the DNN. We examine the efficiency of our method against the augmentation-based systems. For instance, the CamStype-Era and SPGAN used the GAN as the data augmentation, completely fail after the attack of our method. 

\begin{figure}[!t]
\centering
\begin{minipage}[b]{1\linewidth}
\centering
\subfloat[\textbf{The original images.}]{\includegraphics[width=1\textwidth, height=0.23\textwidth]{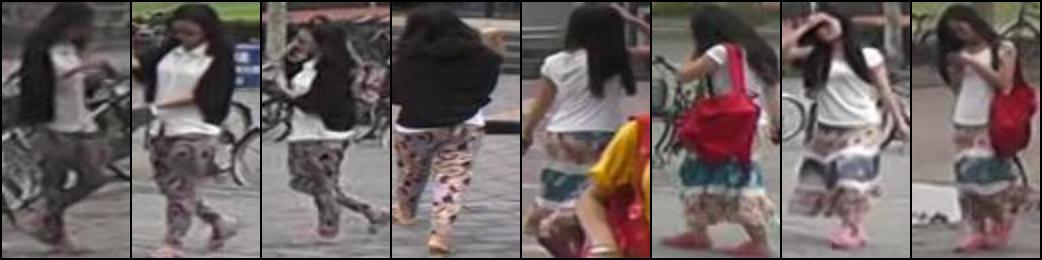}}
\end{minipage}
\begin{minipage}[b]{1\linewidth}
\centering
\subfloat[\textbf{The adversarial examples of DMR\cite{Wang_2020_CVPR}}]{\includegraphics[width=1\textwidth, height=0.23\textwidth]{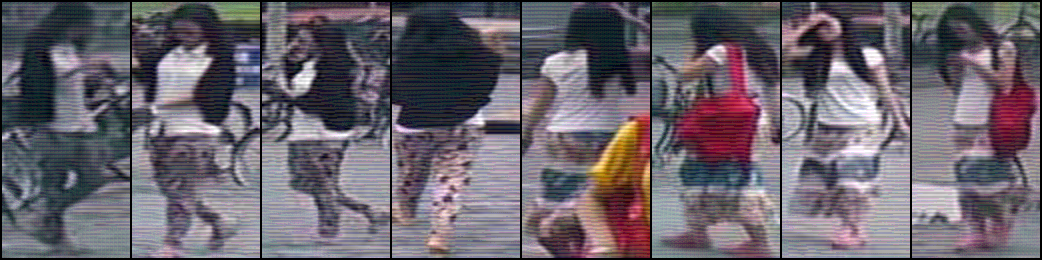}}
\end{minipage}
\begin{minipage}[b]{1\linewidth}
\centering
\subfloat[\textbf{The adversarial examples of Ours.}]{\includegraphics[width=1\textwidth, height=0.23\textwidth]{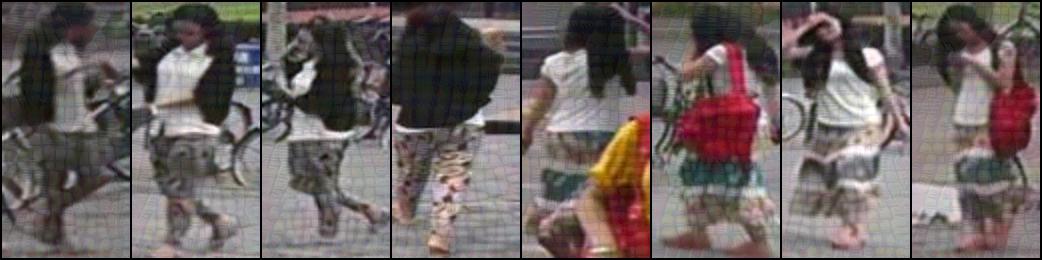}}
\end{minipage}
\caption{The adversarial examples of ours are obviously more deceptive. (\textit{Market1501})}
\label{fig:reid_constract}
\end{figure}

\begin{table}[!t]
\setlength{\abovecaptionskip}{0cm} 
\caption{The experimental results about the transferability of the symmetric saliency-based auto-encoder.}
\centering

\subfloat[\textbf{Cross Models.} The model are trained based on ResNet18 on \textit{imagenette}. The test dataset is \textit{imagenette}. All the accuracy drop sharply in cross model manner.]{
\small
\begin{minipage}[b]{1\linewidth}
\centering
\begin{tabular}{c|cc}
\hline
\hline
\textbf{Tagret Model} &  \textbf{Accuracy} (\%) & \textbf{Degradation} (\%) \\
\hline
ResNet18$\to$EfficientNet-B0 & 48.9 & 42.3$\downarrow$ \\
ResNet18$\to$MobileNetv2  & 17.6 & 76.0$\downarrow$ \\
ResNet18$\to$DenseNet121     & 49.0 &37.2$\downarrow$ \\
ResNet18$\to$GoogleNet    & 31.7 & 58.6$\downarrow$ \\      
\hline
\hline
\end{tabular}
\end{minipage}
\label{tab:cross_model}
}

\subfloat[\textbf{Cross Datasets.} The first two rows are based on Mudeep model, the last two rows are based on PCB model.]{ 
\small
\begin{minipage}[b]{1\linewidth}
\centering
\begin{tabular}{c|c|ll}
\hline
\hline
\textbf{Target Model} & \textbf{Dataset} & \textbf{mAP (\%)} & \textbf{R-1 (\%)}\\
\hline 
\multirow{2}*{Mudeep} & \textit{Market1501}$\to$\textit{cuhk03}      & 15.8(14.3$\downarrow$)       & 26.3(5.8$\downarrow$)  \\
                      & \textit{cuhk03}$\to$\textit{Market1501}      & 3.1(46.8$\downarrow$)        & 3.4(69.6$\downarrow$)   \\
\hline
\multirow{2}*{PCB} & \textit{Market1501}$\to$\textit{cuhk03}      & 14.5(34.1$\downarrow$)        & 18.7(31.9$\downarrow$)  \\
                   & \textit{cuhk03}$\to$\textit{Market1501}      & 20.0(50.7$\downarrow$)       & 35.1(53.5$\downarrow$)  \\
\hline
\hline
\end{tabular}
\end{minipage}
\label{tab:cross_dataset}
}

\subfloat[\textbf{Cross both Datasets and Models.} The left part of the $\to$ is the target model and dataset in the training phase and the right part of the $\to$ is the target model or dataset in the test phase.]{  
\small
\begin{minipage}[b]{1\linewidth}
\centering
\begin{tabular}{c|c|cc}
\hline
\hline
\textbf{Target Model} & \textbf{Dataset} & \textbf{mAP (\%)} & \textbf{R-1 (\%)}         \\ 
\hline
PCB$\to$Mudeep & \textit{Market1501}$\to$\textit{cuhk03} & 12.58 & 19.69 \\
Mudeep$\to$PCB & \textit{cuhk03}$\to$\textit{Market1501}  & 1.13 & 0.98\\
\hline
HACNN$\to$Aligned & \textit{Market1501}$\to$\textit{cuhk03} & 7.81 & 7.98\\
Aligned$\to$HACNN & \textit{cuhk03}$\to$\textit{Market1501} & 35.52 & 45.43\\
\hline 
\hline
\end{tabular}
\end{minipage}
\label{tab:cross_dataset_model}
}

\subfloat[\textbf{Cross Tasks}. The classification task are based on model ResNet18, and on dataset imagenette. The retrieval task are based on model IDE, and on dataset Market1501.]{
\small
\begin{minipage}[b]{1\linewidth}
\centering
\begin{tabular}{c|ccc}
\hline
\hline
\textbf{Task} & \textbf{Accuracy (\%)} & \textbf{mAP (\%)}\\
\hline
Classification$\to$Retrieval & - &63.30$\to$38.57 \\
Retrieval$\to$Classification &  93.96$\to$43.11 & - \\
\hline
\hline

\end{tabular}
\end{minipage}
\label{tab:cross_task}
}

\end{table}
\setlength{\abovecaptionskip}{0cm} 

\noindent \textbf{Visual Quality.} From Table \ref{tab:ir}, we can find that the DMR can also successfully attack the corresponding models and outperforms our method in some cases. However, the most important advantage of our method lies in the visual quality of the adversarial examples. We get this conclusion from two aspects: (1) the image quality assessment measures ssim/ms-ssim/psnr of our method are much higher than those of the DMR. (2) We visualize the adversarial examples in Figure \ref{fig:reid_constract} and find that the perturbations of the DMR are much more obvious than ours. 


\begin{figure*}[!t]
\begin{minipage}[b]{0.45\linewidth}
\subfloat[Image Classification (ResNet18, imagenette)]{\includegraphics[width=1\textwidth, height=1.4\textwidth]{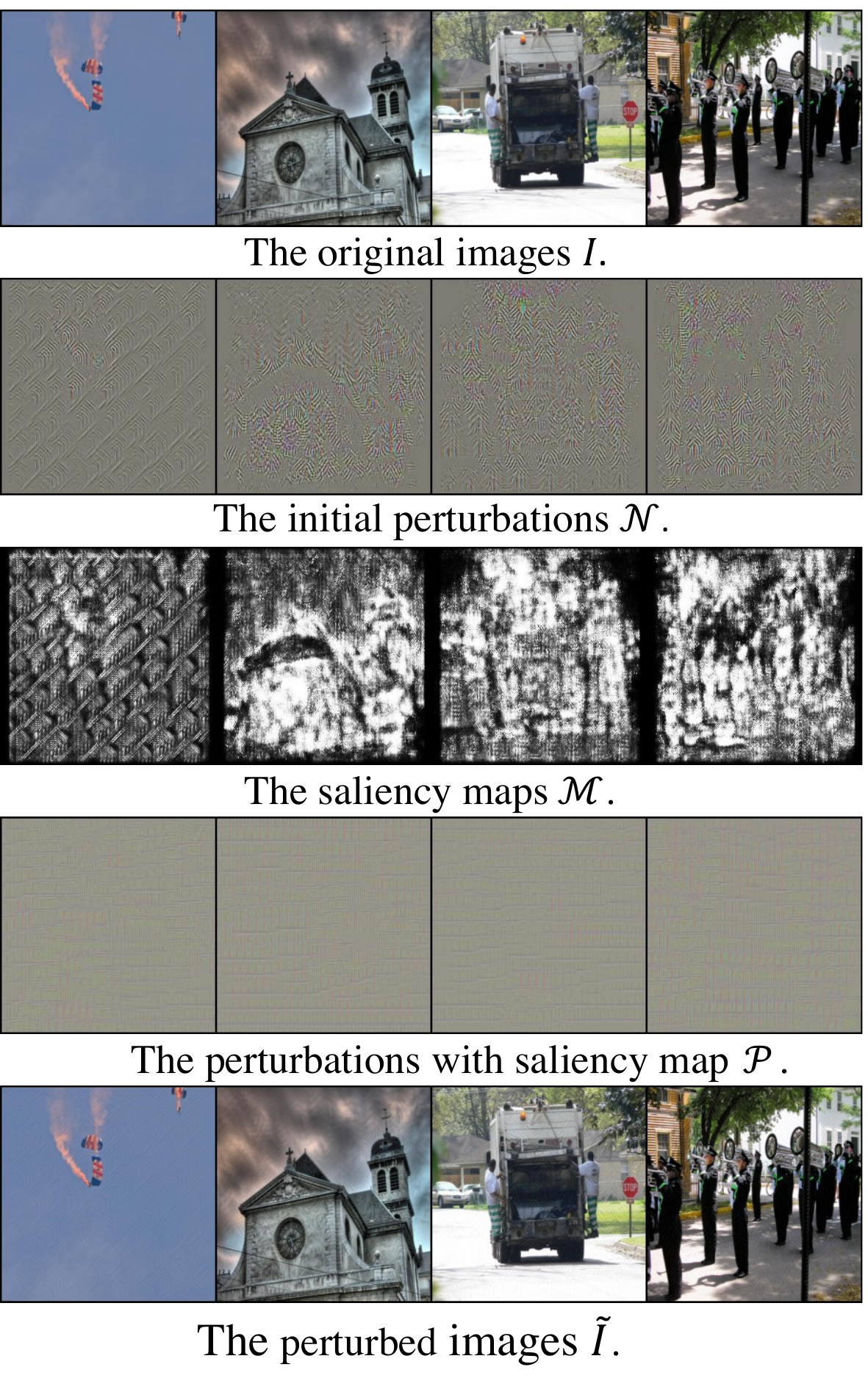}}
\end{minipage}
\begin{minipage}[b]{0.45\linewidth}
\subfloat[Image Retrieval (AlignedReID, Market1501)]{\includegraphics[width=1\textwidth, height=1.4\textwidth]{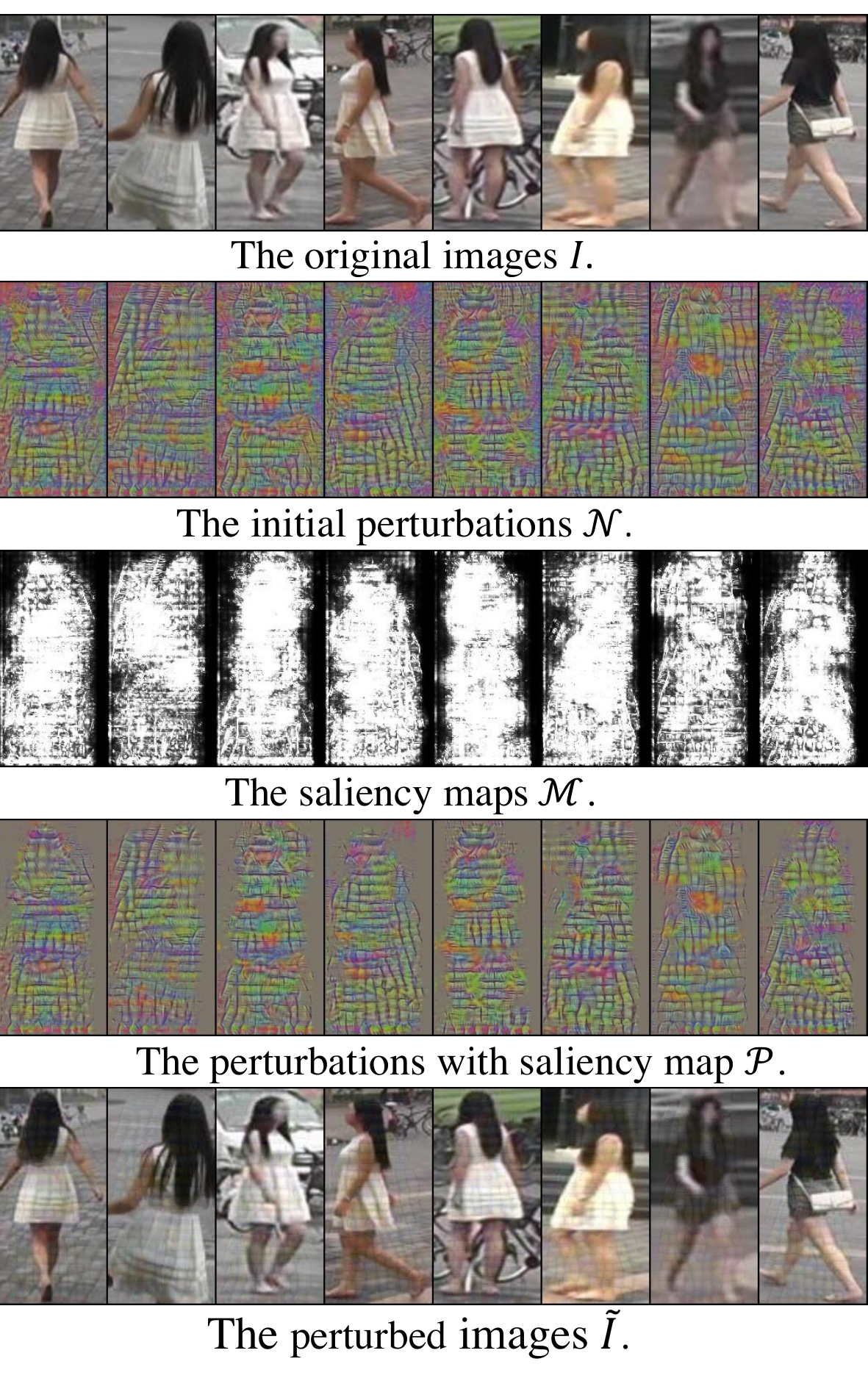}}
\end{minipage}
\caption{The visualization of perturbation maps and saliency maps. The white regions in saliency maps mean high label-relevant, while the black regions mean low label-relevant. The perturbations with saliency map are obviously more deceptive. (Best viewed with zoom in.)}
\label{fig:ablation}
\end{figure*}

\subsection{Transferability} 
The transferability is an important metric for adversarial attack because in the real-world scene we always have no knowledge about the target model. One of the benefits of generative-based attacks comes from their transferability. We evaluate the transferability of our method from four aspects: (1) cross models, (2) cross datasets (3) cross both models and datasets and (4) cross tasks. 

\noindent \textbf{Cross Models.} We convey the cross-model experiments in classification tasks. We train the SSAE in attacking ResNet18 and test its performances on attacking other models, including the EfficientNet-B0, the MobileNetv2, the DenseNet121 and the GoogLeNet, and obtain the results in Table \ref{tab:cross_model}. We can find that our attack gets good attack ability in the MobileNetv2 and the GoogLeNet, whose accuracy drop about 76.0\% and 58.7\%, respectively. When attacking the EfficientNet-B0 and the 
DenseNet, the classification systems can retain about 50\% accuracy, and the attack ability is relatively low. However, in real-world applications, we think 50\% accuracy already means that the systems break down.

\noindent \textbf{Cross Datasets.} We convey the cross-dataset experiments of image retrieval task in Table \ref{tab:cross_dataset}. We use the Mudeep and the PCB as our target models. All the mAP and Rank-1 have certain degradation, which means our method can work in a cross-dataset manner.

\noindent \textbf{Cross both Datasets and Models}. We convey the cross-dataset and cross-model experiments in the person re-identification task and obtain the results in Table \ref{tab:cross_dataset_model}. We find that the attack ability on the cross-dataset and cross-model setting are even better than the cross-dataset setting only. 
The cross models and the cross datasets setting can be viewed as semi-black box and the cross dataset and corss-model can be viewed as black-box settings. That is to say, The experimental results validate the effectiveness of SSAE in a black-box manner on the one hand.

\noindent \textbf{Cross Tasks}. Actually, the cross-task setting is not practical in application, because we always know the task of a real-world system in advance. We test this manner for further validating that our method can be easily transferred to another task. We use the SSAE that trained in the image classification task to perturb the instances in the person re-identification task, vice versa, and obtain the results in Table \ref{tab:cross_task}. The results validate that our attack is transferable even in the cross-tasks scene.

\subsection{Consumption of Resources}
We have mentioned that the generative-based methods are resource-saving compared to the gradient-based method. To further illustrate the benefits of the generative-based attack method, we conduct the corresponding experiments on the speed (FPS) and the load on GPU memory (MiB). We test the speed and the load on GPU memory on PyTorch, one GTX1080 with batch size as 1. We use a public library, AdverTorch \cite{ding2019advertorch}, to reproduce the performance of PGD and FGSM. Table \ref{tab:consumption} shows that the generative-based methods are both more time-saving and GPU-memory saving comparing with the gradient-based methods (i.e., FGSM, PGD). The PGD needs multiple times of gradient backpropagation to calculate the perturbations, thus leading to poor performance of speed. Although the FGSM only needs one time of gradient backpropagation, its speed depends on the architecture of the target model, while our proposed method is based on a lightweight auto-encoder with only forward inference once. Both the PGD and the FSGM calculate and save the gradients, thus the load on GPU memory of them is much higher than generative-based methods.


\subsection{Implementation Details}
\noindent \textbf{The setting of training target models.} All the classification models in the experiments are trained by ourselves. All the training settings are the same amount three dataset, except the input size of \textit{cifar-10} is $32\times32$ instead of $224\times224$. All the target models in person re-id are provided by Wang et.al \cite{Wang_2020_CVPR}. 

\noindent \textbf{The setting of training the auto-encoder.}
We train SSAE with $\mathcal{L}_{angular}$ in Eq.(\ref{eq:cos}) only for the first 20 epochs and then with the $\mathcal{L}$ in Eq.(\ref{eq:objective}) for another 20 epochs. We train $\mathcal{L}_{angular}$ alone at the beginning because the loss function $L_{f}$ tends to dominate the optimization if we train loss $L$ directly. The batch size in our experiments is 16 for both image classification and person re-identification. We use Adam as the optimizer and set the learning rate as 1e-5. The hyperparameter $\alpha$ in Eq.(\ref{eq:objective}) is set as 0.0001 in our experiments. 

\subsection{Effect of Saliency map}
Although the saliency map module does not use any label supervision directly, it can make the model pay more attention to the label-relevant region with angle-norm disentanglement module together. We randomly choose some examples, and visualize their corresponding perturbations, the saliency map and adversarial examples, respectively in Figure \ref{fig:ablation}. From the saliency maps $\mathcal{M}$ in the third row of Figure \ref{fig:ablation}, we find that the values of the label-relevant regions are closed to 1, while the values of the regions of the label-irrelevant regions are closed to 0. The advantage compared to the binary mask (e.g., Mask-RCNN) is that the saliency map can reflect the relative importance of different regions and make the perturbation smoother. The visual results are in line with the original design intention of the saliency map module. And the saliency map module not only works on the images with one object but also can be extended to the images with multiple objects: the fourth image in \ref{fig:ablation.a} has numerous objects and the corresponding saliency map can reflect it. Applying the saliency maps to the initial perturbations makes them more inconspicuous.



\section{Conclusion}

This work proposes a discriminator-free generative-based adversarial attack method. To improve the visual quality, a saliency map module is used to limit the added perturbations within a sub-region of the whole image. 
We disentangle the norm and the angle of the CNN features, push away the angles between raw instances and perturb instances to make the CNN model failed. The abundant experiments validate the effectiveness of our proposed attack, even in the transferable manner.
\newpage
\bibliographystyle{ACM-Reference-Format}
\bibliography{sample-base}

\end{document}